# Design a Persian Automated Plagiarism Detector (AMZPPD)

Maryam Mahmoodi [#1], Mohammad Mahmoodi Varnamkhasti [*2]

[1] *Department of computer, Meymeh branch, Islamic Azad University, Meymeh, Iran*
[2] *Department of computer, University of Isfahan, Isfahan, Iran*

*Abstract*— **Currently there are lots of plagiarism detection approaches. But few of them implemented and adapted for Persian languages. In this paper, our work on designing and implementation of a plagiarism detection system based on pre-processing and NLP technics will be described. And the results of testing on a corpus will be presented.**

## I. INTRODUCTION

Todays, internet made sharing documents and information very easy. And consequently written-text plagiarism is now a real problem for most academic environments.

Many methods and approaches have been developed for automatic detection of plagiarism mostly for English language. But for Persian language there are few works. So in this work the aim is to implement an accurate plagiarism detection system with the purpose of short paragraphs.

## II. ARTICHETURE OF METHODOLOGY

In this section we described our system methodology that detects how similar are two suspicious texts.

In [1], a multi–stage new framework introduced that is adapted to Persian language in our work and used as the main Idea:

### A. Pre-Processing Stage

In this stage some pre-processing techniques are used to extract single words from the structured text and removal of unnecessary things that make it difficult to recognize similarities. In section III, techniques and their techniques and the importance of each is described in Persian.

### B. Similarity comparison Stage

In this stage one of these comparison methodologies is used to compare two sequences of words were produced in the previous stage:

  2-gram similarity measures (using Jaccard or Clough & Stevenson used similarity metric)

  3-gram similarity measures(using Jaccard or Clough & Stevenson used similarity metric)

  Longest common subsequence

### C. Verdict stage

Using numbers generated in pervious stages one of these verdicts is derived:

Clean (Non-plagiarism): based on participants own knowledge as the original texts were not given.

Heavy revision: rewriting of the original by paraphrasing and restructuring.

Light revision: minor alteration of the original text by substituting words with synonyms and performing some grammatical changes.

Near copy:  copy-and-paste from the original text.
[1]

This may be obtained by single-criteria or multi-criteria analysis.

## III. PRE-PROSECCING

### A. Normalizing

A Persian Normalizer removes extra spaces, organizes virtual spaces (for example correct "می روم" to "می‌روم"), fix the problem of "ي" and "ى". This pre-processing technic is very important in Persian because most texts do not adhere to correct orthography. And it may lead to difficulty in detecting the similarity.

### B. Stop-word Removal

This technique removes common words (articles, prepositions, determiners,…) such as "از","به","که". These words sometimes have very little influence in meaning but sometimes play somewhat important role in the text.

Two kind of deep and shallow stop-word-removal tables tested to find the true impact of this technic.

### C. Sentence segmentation

Split text in the document into sentences and thereby allowing line-by-line processing in the subsequent tests [1].

### D. Tokenization

Token (words, punctuation symbols, etc.) boundaries in sentences[1].

*E. Stemming*

Stemming is the process of removing and replacing word suffixes to arrive at a common root form of the word. Sometimes people change the form of a word to hide their plagiarism.

Suppose these two sentences:

"این پردازنده ها می توانند پردازش سیگنال را انجام دهند"

And

"این نوع پردازنده توان پردازش سیگنال را دارند"

In this case the similarity checker will make mistake if words don't change into stems.

*F. Lemmatization*

Transform words into their dictionary base forms in order to generalize the comparison analysis [1]. Sometimes lemmatization is mistaken for stemming. However, there is an essential difference. Stemming operates only with single words without any knowledge of the context, and therefore cannot distinguish among words having several different meanings [2].

*G. Number Replacement*

This one replaces any number with a dummy character. ("#" for example) The reason for doing this is that in some scientific reports, dishonest person can just change the numbers carefully to cheat the system.

*H. Synonymy Recognition*

The motivation for using synonymy recognition comes from considering human behaviour, whereby people may seek to hide plagiarism by replacing words with appropriate synonyms [2].

In Persian Language synonymy recognition is very important because any word have many synonyms. And sometimes synonyms are the foreign equivalent of the words in Persian script. For example "کیبورد" can be used instead of "صفحه کلید" In order to mislead the plagiarism detector.

*I. Part-of-Speech tagging*

Assign grammatical tags to each word, such as "noun", "verb", etc., for detecting cases where words have been replaced, but the style in terms of grammatical categories remains similar[1].

IV. SIMILARITY COMPARISON

With the end of the first stage, we have two sequences of words. One of the original text and one of the suspicious text.

These sequences where cleaned by some of the techniques mentioned above and are ready to check by one of these similarity comparison methods:

*A. N-grams + Jaccard similarity coefficient*

If S(A) is the set of n-grams of original text and S(A) is the set of n-grams of suspicious text. The Jaccard similarity coefficient is defined as:

$$J(A,B) = \frac{|S(A) \cap S(B)|}{|S(A) \cup S(B)|}$$

*B. N-grams + Clough & Stevenson metric*

In similar conditions with part A Clough & Stevenson metric defined as [1, 3]:

$$C(A,B) = \frac{|S(A) \cap S(B)|}{|S(A)|}$$

*C. LCS*

The longest sequence of words included in both original and suspicious documents may be used as the criteria of similarity.

V. TEST COMBINATIONS

In III and IV some candidates for pre-processing and similarity comparison were introduced.

To identify the best possible combination, we tested most possible ones in table1 and table2 are the information of top10 combiantions.

For testing different combinations we first developed a Persian Corpus just like [3] that is available at:

http://amzmohammad.com/AMZPPD/CPPD.tar

TABLE I
AVERAGE RESULTS OF SAMPLE COMBINATIONS

| Combiantion | Average similarity score | | | |
|---|---|---|---|---|
| | Clean | Heavy Revision | Light Revision | Copy |
| 1 | 0.0019 | 0.003 | 0.039 | 0.198 |
| 2 | 0.004 | 0.016 | 0.112 | 0.294 |
| 3 | 0.002 | 0.003 | 0.038 | 0.223 |
| 4 | 0.021 | 0.037 | 0.124 | 0.314 |
| 5 | 0.0043 | 0.007 | 0.075 | 0.248 |
| 6 | 0.0022 | 0.035 | 0.199 | 0.403 |
| 7 | 0.0044 | 0.017 | 0.077 | 0.302 |
| 8 | 0.0052 | 0.070 | 0.224 | 0.438 |
| 9 | 0.207 | 0.230 | 0.245 | 0.267 |
| 10 | 0.18 | 0.187 | 0.191 | 0.219 |

TABLE II
COMBINATION DESCRIPTIONS

| Combination | Normalizing | Stop-word | Sentence segmentation | tokenization | Stemming | Lemmatization | Number replacement | Synonymy recognition | Part-of-speech tagging | Similarity compare |
|---|---|---|---|---|---|---|---|---|---|---|
| 1 | Y | D | Y | Y | Y | Y | Y | Y | Y | 3-gram+jaccard |
| 2 | Y | D | Y | Y | Y | Y | Y | Y | N | 2-gram+jaccard |
| 3 | Y | S | Y | Y | Y | Y | N | Y | Y | 3-gram+jaccard |
| 4 | Y | S | Y | Y | Y | Y | Y | Y | Y | 2-gram+jaccard |
| 5 | Y | D | Y | Y | N | Y | Y | Y | N | 3-gram+Clough |
| 6 | Y | D | Y | Y | Y | Y | Y | Y | N | 2-gram+Clough |
| 7 | Y | S | Y | Y | N | Y | Y | Y | N | 3-gram+Clough |
| 8 | Y | S | Y | Y | Y | Y | N | Y | N | 2-gram+Clough |
| 9 | Y | D | Y | Y | N | Y | Y | Y | N | LCS |
| 10 | Y | D | Y | Y | Y | Y | Y | Y | Y | LCS |

These ten combinations are selected because had meaningful difference in values obtained in the different groups.

VI. CHOOSE THE BEST COMBINATION

The criterion is coefficient of dispersion. That is defined as the ratio of the variance to the mean in each group (verdict),

$$D = \frac{\sigma^2}{\mu}.$$

VII. CONCLUSIONS

In this work we had shown that the influence of NLP technics and pre-proceedings on Persian plagiarism detection accuracy is Significant. But because of orthography problems and maybe the ambiguities of language this influence is less than English.

The result of this work is now implemented using python, NLTK and HAZM library and is under testing and development in some academic environments.

Under the project name AMZPPD: http://amzmohammad.com/AMZPPD/

A good combination has a little coefficient of dispersion in all verdicts. Because if coefficient of dispersion is a big number we cannot determine accurate intervals of verdicts and they may have overlap.

As the resultant testing combinations on corpus, the four combinations 5,6,8,9 have better coefficient of dispersion and are our chosen combinations.